\def\year{2022}\relax
\documentclass[letterpaper]{article} % DO NOT CHANGE THIS
\usepackage{aaai22}  % DO NOT CHANGE THIS
\usepackage{times}  % DO NOT CHANGE THIS
\usepackage{helvet} % DO NOT CHANGE THIS
\usepackage{courier}  % DO NOT CHANGE THIS
\usepackage[hyphens]{url}  % DO NOT CHANGE THIS
\usepackage{graphicx} % DO NOT CHANGE THIS
\usepackage{natbib}  % DO NOT CHANGE THIS AND DO NOT ADD ANY OPTIONS TO IT
\usepackage{caption} % DO NOT CHANGE THIS AND DO NOT ADD ANY OPTIONS TO IT
\usepackage{amssymb}
\usepackage{bm}
\usepackage[table,x11names]{xcolor}
\usepackage[utf8]{inputenc}
\usepackage[english]{babel}
\usepackage{multirow}
\newcommand{\specialcell}[2][c]{%
  \begin{tabular}[#1]{@{}c@{}}#2\end{tabular}}
\usepackage{makecell}
\usepackage{booktabs} % For formal tables
\usepackage{subcaption}
\usepackage{amsmath}
\usepackage{enumitem}
\usepackage{algorithm}
\usepackage{algorithmicx}
\usepackage{algpseudocode}
\usepackage{footnote}
\makesavenoteenv{tabular}
\newtheorem{assumption}{Assumption}
\usepackage{dsfont}
\definecolor{ForestGreen}{rgb}{0.0, 0.27, 0.13}
\newcommand{\mat}[1]{{\bf #1}}   % matrix: bold
\newcommand\independent{\protect\mathpalette{\protect\independenT}{\perp}}
\def\independenT#1#2{\mathrel{\rlap{$#1#2$}\mkern2mu{#1#2}}}
\title{Effects of Multi-Aspect Online Reviews with Unobserved Confounders: Estimation and Implication}
\author{
Lu Cheng\textsuperscript{\rm 1}, Ruocheng Guo\textsuperscript{\rm 2}, Kasim Candan\textsuperscript{\rm 1}, Huan Liu\textsuperscript{\rm 1}\\
}
\affiliations{
\textsuperscript{\rm 1} School of Computing and Augmented Intelligence, Arizona State University, USA\\
\textsuperscript{\rm 2} School of Data Science, City University of Hong Kong, China\\
\{lcheng35, candan, huanliu\}@asu.edu, ruocheng.guo@cityu.edu.hk % email address must be in roman text type, not monospace or sans serif
}

\graphicspath{{figures/}}
\begin{document}

\maketitle

\begin{abstract}
Online review systems are the primary means through which many businesses seek to build the brand and spread their messages. Prior research studying the effects of online reviews has been mainly focused on a single numerical cause, e.g., ratings or sentiment scores. We argue that such notions of causes entail three key limitations: they solely consider the effects of single numerical causes and ignore different effects of multiple aspects -- e.g., Food, Service -- embedded in the textual reviews; they assume the absence of hidden confounders in observational studies, e.g., consumers' personal preferences; and they overlook the indirect effects of numerical causes that can potentially cancel out the effect of textual reviews on business revenue. We thereby propose an alternative perspective to this single-cause-based effect estimation of online reviews: \textit{in the presence of hidden confounders}, we consider \textit{multi-aspect} textual reviews, particularly, their total effects on business revenue and direct effects with the numerical cause -- ratings -- being the mediator. We draw on recent advances in machine learning and causal inference to together estimate the hidden confounders and causal effects. We present empirical evaluations using real-world examples to discuss the importance and implications of differentiating the multi-aspect effects in strategizing business operations.
\end{abstract}

\section{Introduction}
The low cost of gathering and distributing information in online review systems has greatly facilitated a large-scale of crowd-sourced reviews via the electronic Word of Mouth. Prior research has established the importance of studying effects of online reviews in guiding consumer choices. For instance, positive reviews and popularity of reviews can largely influence book sales \cite{chevalier2006effect} and restaurant reservation availability \cite{anderson2012learning}. Many of the leading notions of causes in these studies are single numerical causes\footnote{We use the ``cause'' to represent the conventional ``treatment''. As some treatments may not exhibit causal effects, a more precise term would be ``potential causes''. We use ``cause'' for simplicity.} such as a numerical rating of a restaurant or an aggregated sentiment score of a textual review. Despite its simplicity, this approach cannot provide a granular-level analysis of existing problems in businesses, resulting in its limited use and coverage in practice \cite{sachdeva2020useful}. Online reviews typically encompass rich contextual information, e.g., content in the textual reviews, beyond the simple statistics such as ratings. We argue that current works using single numerical causes comes with three limitations:
\begin{figure}
\centering
  \includegraphics[width=\linewidth]{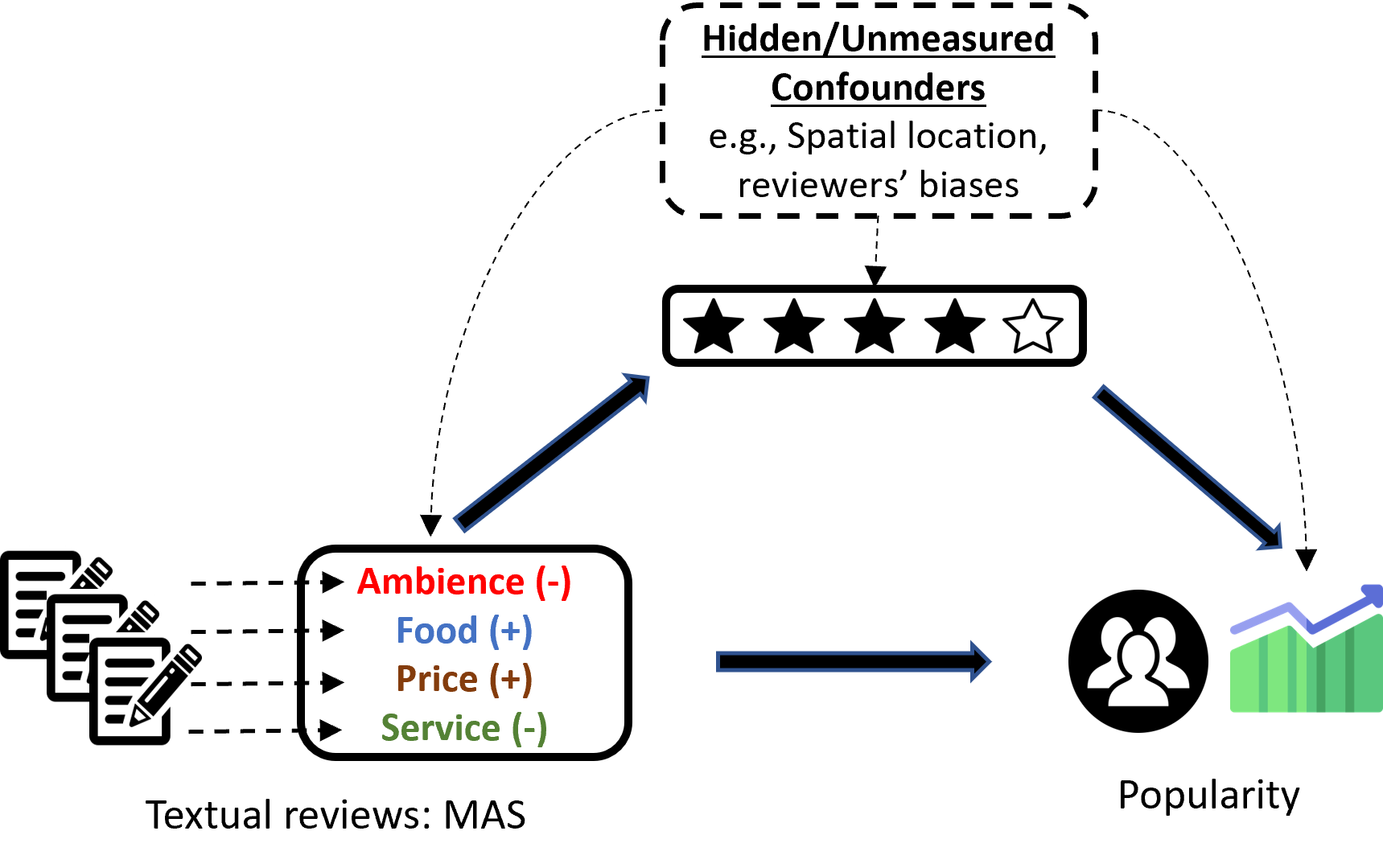}
  \caption{Problem illustration with causal diagram. Given MAS extracted from a corpus of textual reviews, ratings, and popularity, we examine, \textit{in the presence of hidden confounders} (the dashed rectangle): 1) how \textit{MAS} (potential causes) influences \textit{ratings} (outcome); 2) how \textit{MAS} influences \textit{restaurant popularity} (outcome); and 3) how \textit{MAS} directly influences \textit{popularity} while being mediated by \textit{ratings} (mediator).}
 \label{problem}
\end{figure}
\begin{itemize}[leftmargin=*]
    \item Reviews often describe multiple potential aspects~\cite{fan2018multi}, and each aspect provides a unique assessment. For example, the following review (color-coded) with an overall rating of 3 stars from Yelp.com\footnote{https://www.yelp.com/} expresses different sentiments toward multiple aspects (i.e., multi-aspect sentiment) -- a positive sentiment toward the restaurant's food and negative opinions toward its ambience and service: ``\textit{The sushi was very good (food), but it took over half an hour to be seated (service). The room was very noisy and cold, wind blew in from a curtain next to our table (ambience)}''. 
    \item Most works assume the absence of hidden confounders, unobserved/unmeasured variables that cause spurious associations between the outcome (e.g., restaurant popularity) and causes (e.g., multi-aspect sentiment scores) \cite{rubin1976inference,pearl2009causality,guo2020survey,yao2020survey}. The assumption is, however, unverifiable in practice. For instance, consumers' personal preferences can simultaneously confound the sentiment aspects and restaurant popularity but are often unobserved/unmeasured. When left out, such confounding bias can lead to inaccurate and inconsistent causal effect estimation \cite{rosenbaum1984reducing,pearl2009causality,wang2019blessings}.   
    \item A typical online review consists of a numerical rating and a chunk of review text. Due to their similar functionality, textual reviews and numerical ratings (i.e., the mediator between the textual reviews and outcome) might compete with each other influencing the outcome of interest (e.g., business popularity), as illustrated in Figure \ref{problem}. The effects of textual reviews, therefore, can be cancelled out (i.e., become less significant) by ratings. Nevertheless, few related discussions have been observed in the field. 
\end{itemize}

To address these limitations, we provide an alternative perspective to the single-cause-based effect estimation of online reviews. Particularly, \textit{in the presence of hidden confounders}, we study the effects of multi-aspect sentiment (MAS) identified in the textual reviews: both the total effects\footnote{Total effect = direct effect + indirect effect.} and direct effects with ratings being the mediator. As described in Figure \ref{problem}, we discuss three types of causal relations among ratings, MAS, and business popularity: (1) \textit{MAS} $\rightarrow$ \textit{ratings}; (2) \textit{MAS} $\rightarrow$ \textit{popularity}; and (3) \textit{MAS} $\rightarrow$ \textit{ratings} $\rightarrow$ \textit{popularity}. We are particularly interested in the outcomes related to the business revenue, namely, restaurant popularity, defined as the average hourly consumer flow within a specific day. We further propose a principled framework that combines techniques in machine learning and causal inference to estimate the effects while accounting for hidden confounders. We follow the causal mechanism illustrated in Figure \ref{problem} and ask the following research questions:
\begin{itemize}[leftmargin=*]
    \item \textbf{RQ. 1} How does our framework differ from non-causal methods w.r.t. prediction and effects estimation results? 
    \item \textbf{RQ. 2} Which sentiment aspects have causal effects on the restaurant \textit{ratings} and how different are these effects?
    \item \textbf{RQ. 3} Which sentiment aspects are causally related to the restaurant \textit{popularity} and how different are these effects?
    \item \textbf{RQ. 4} Can MAS provide additional information about restaurant popularity besides ratings?
\end{itemize}
\textbf{RQ. 1} provides empirical evaluations to illuminate the validity and efficacy of our framework alleviating confounding bias in observational studies. According to Figure \ref{problem}, \textbf{RQ. 2-3} seek to examine the causal effects of MAS on ratings and the total effects on popularity. \textbf{RQ. 4} further investigates the direct effects of MAS on restaurant popularity with ratings being the mediator. 

\noindent\textbf{Contributions.} With the consideration of hidden confounders, we propose to investigate causal effects of textual reviews from multiple dimensions in order to identify aspects most relevant to business revenue. Our first contribution sheds light on the importance of differentiating multi-aspect effects in strategizing business operations. As with other observational studies, a major challenge in this work is to control for hidden confounders that might render biased and inconsistent effect estimations. Drawing on recent advances in machine learning and causal inference, our second contribution is a principled framework that infers \textit{surrogate confounders} from MAS to control for hidden confounders. Lastly, we conduct extensive evaluations on novel datasets curated by combining two independent data sources -- Yelp and Google Map\footnote{https://maps.google.com/}, and discuss practical implications.
\section{Related Work}
\textbf{Multi-Aspect Sentiment Analysis.} Aspect-level sentiment analysis \cite{kumar2016opinion} is conventionally regarded as a text classification task where informative features are extracted to train a multi-class classifier. For example, Lu et al.~\cite{lu2011multi} proposed a weakly-supervised approach that leveraged seed words as prior knowledge to enforce a direct connection between aspect and seed words. Vo and Zhang \cite{vo2015target} designed a sentiment-specific word embedding and sentiment lexicons to enrich the input features for prediction. Highly dependent on input features, these models have been gradually replaced by neural-network-based approaches such as recursive neural network \cite{dong2014adaptive}, LSTM \cite{tang2015effective}, and attention-mechanism-based models \cite{fan2018multi}. \\
\textbf{Causal Inference with Multiple Treatments.} One of the most common techniques used in causal effect estimation with multiple treatments is generalized propensity scores (GPS) \cite{austin2018assessing}, an extension of propensity score with binary treatment. GPS has been increasingly used in standard causal inference models such as inverse probability of treatment weighting \cite{mccaffrey2013tutorial}, matching \cite{dehejia2002propensity}, subclassification \cite{rosenbaum1984reducing} and imputations \cite{gutman2015estimation}. These approaches simply assume the absence of hidden confounders that typically persist in the observational studies. In computational genetics, a variety of methods have been proposed to account for hidden confounders, e.g., \cite{song2015testing}. The growing interest of controlling hidden confounders can be also found in the field of computer science. More recently, a new approach for multiple effect estimation with hidden confounders combined techniques in unsupervised learning and theories in causal inference to provably eliminate confounding biases \cite{wang2019blessings}.\\
\textbf{Causal Effect Estimation in Online Review Systems.} Various research fields, such as marketing science and economy, have shown increasing interest in the effects of online reviews. The outcome of interests spans from sales to competition and consumer welfare \cite{Fang2019}. For example, findings from \cite{chevalier2006effect} suggested a positive relationship between ratings and book sales. A similar study \cite{zhu2010impact} investigated the effect of product features and consumer characteristics from online reviews on sales. In contrast to the positive effects, researchers also examined how manipulating the display design of online review systems can greatly influence restaurant revenue \cite{luca2016reviews}. Conclusions drawing on regression discontinuity design \cite{thistlethwaite1960regression} manifested that an increase in displayed ratings by one star raises the revenues of independent restaurants by 5\%-9\% \cite{luca2016reviews}.

Informed by the three lines of research, this work argues for a more holistic understanding of the effects of online review systems on business revenue. We seek to differentiate the consumer evaluations w.r.t each business aspect and discuss how multi-aspect textual reviews and numerical ratings influence business operations simultaneously. A granular analysis of textual reviews can help identify problems in existing business in detail. Central to our framework is the intersection of machine learning and causal inference to jointly estimate hidden confounders and causal effects. This new perspective is not intended to entirely solve the concerns in estimating effects of online review systems, but rather to elucidate them and bring to the forefront concerns that have been neglected in literature. 
\section{Data}
We follow a similar data collection process described in \cite{luca2016reviews} and curate two novel datasets that merge independent data sources for online reviews and restaurant popularity, respectively. The first data source is the Yelp.com, a platform that publishes crowd-sourced reviews about businesses. When consumers searches Yelp.com, Yelp presents them with a list of businesses that meet their search criteria. Businesses are ranked according to the relevance and ratings, and for each business, the contact information and a short excerpt from one review are also displayed. To access to the entire history of reviews for that business, one needs to click on the specific business. The Yelp dataset\footnote{https://www.yelp.com/dataset/challenge} presents information about local businesses in ten metropolitan areas across two countries (U.S. and Canada). We filtered out non-restaurant businesses based on their category descriptions on Yelp and selected the top two cities with the largest sample sizes: Las Vegas, U.S. and Toronto, Canada. We refer to these two datasets as \textit{LV} and \textit{Toronto}.
\begin{table}
\begin{center}
\resizebox{\columnwidth}{!}{
\begin{tabular}{ c|c|c|c } \hline
Dataset & Sample Size &Ratings (Range) &\#Reviews (Range)\\ \hline
\textit{LV}&3,041 &3.47 (1--5) & 255 (3--8,570) \\ \hline
\textit{Toronto}&3,828&3.50 (1--5)& 67 (3--2,177)\\ \hline
\end{tabular}}
\caption{Dataset statistics of Yelp reviews. Data in the last two columns denote the mean values per restaurant.}
\label{review}
\end{center}
\end{table}

\begin{figure}
\centering
\begin{subfigure}{0.8\columnwidth}
\centering
  \includegraphics[width=\linewidth]{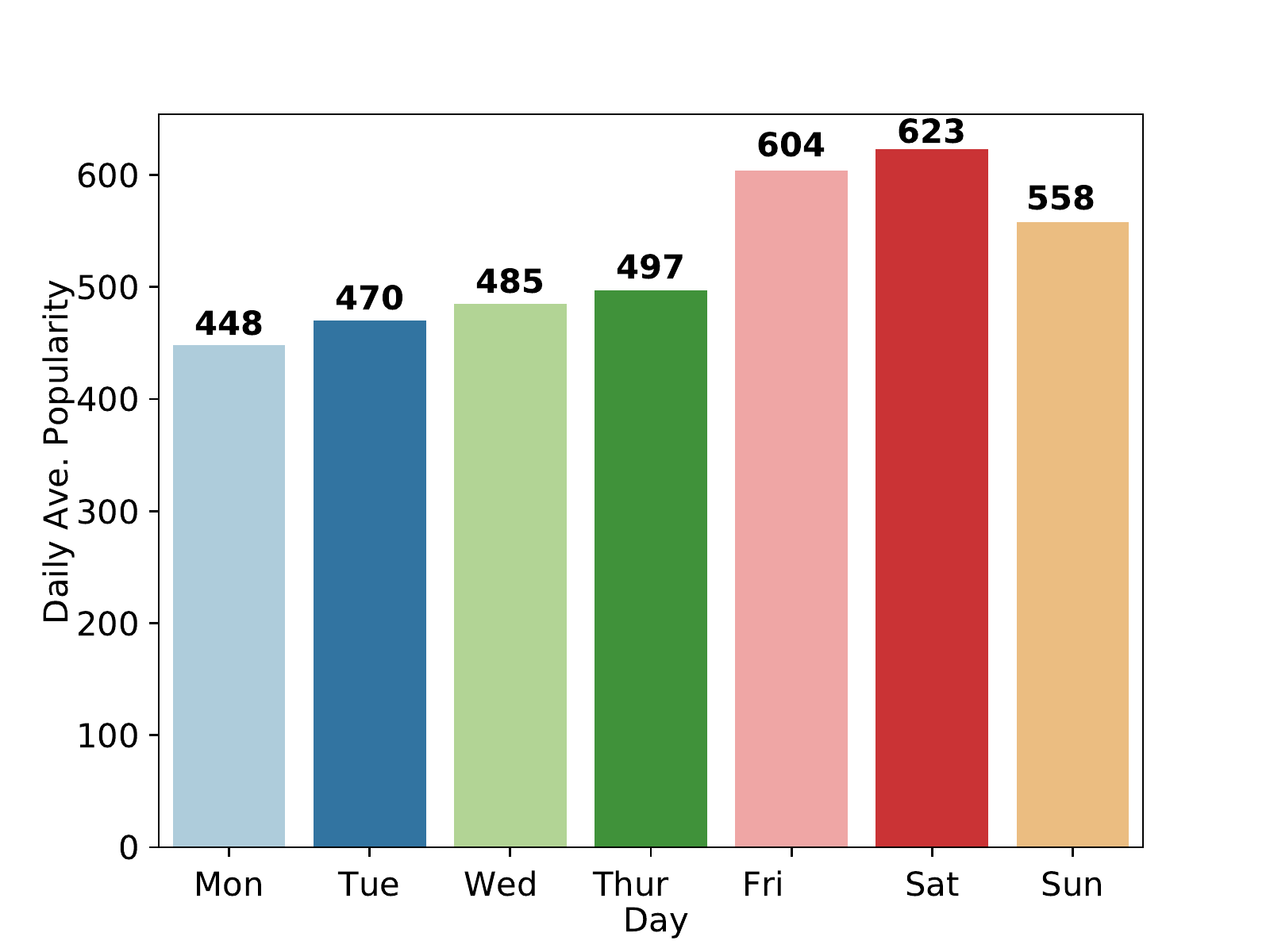}
  \caption{\textit{LV} data.}
\end{subfigure}
\begin{subfigure}{0.8\columnwidth}
\centering
  \includegraphics[width=\linewidth]{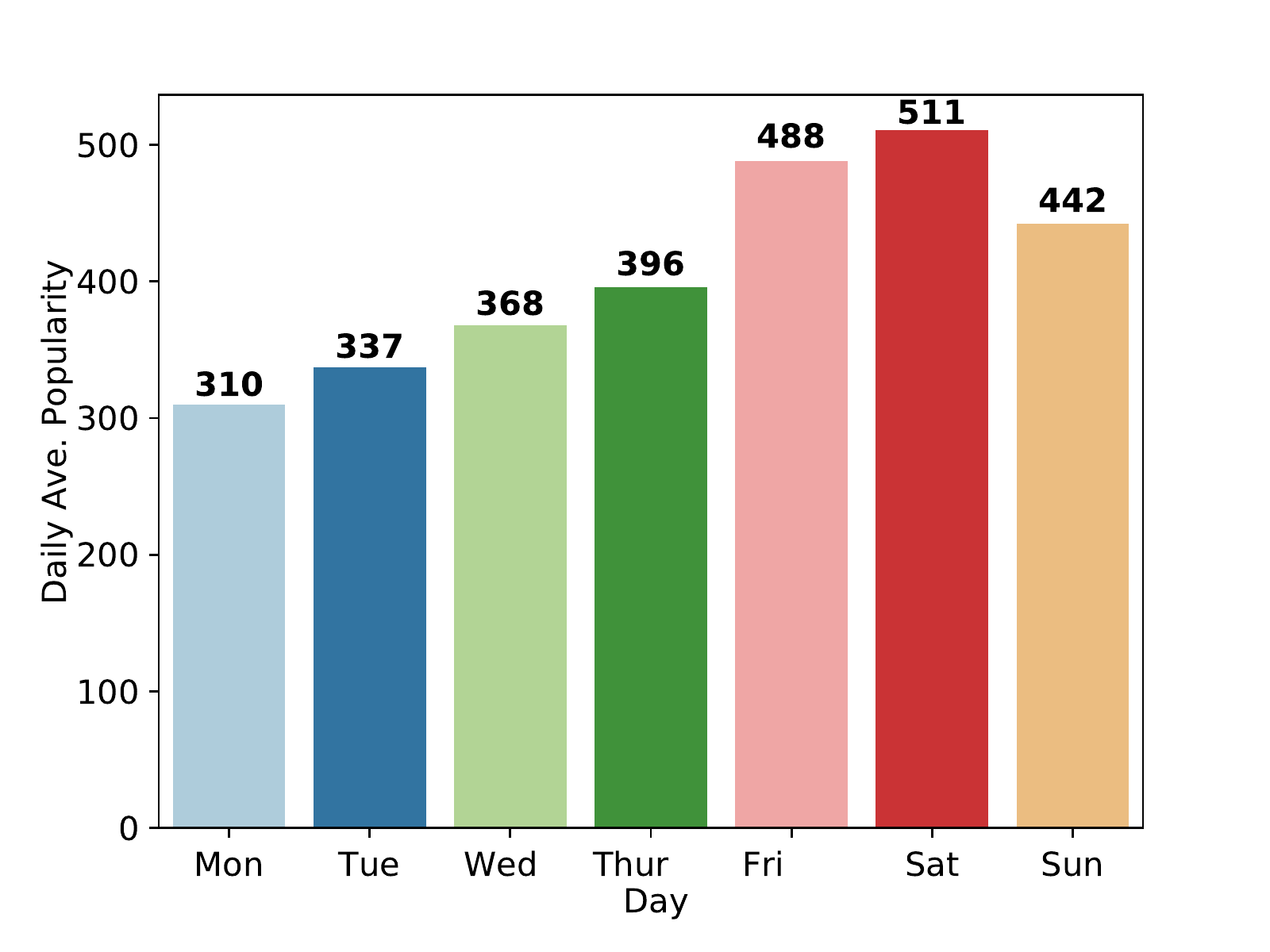}
  \caption{\textit{Toronto} data.}
\end{subfigure}
\caption{Daily average popularity of restaurants over a week.}
 \label{daily}
\end{figure}

The second data source for restaurant popularity comes from Google Map. Particularly, we used Google Popular Times\footnote{https://support.google.com/business/answer/6263531?hl=en} that features restaurant hourly popularity from Monday to Sunday as a surrogate. Popular times measure real-time consumer flow using the Global Positioning System. Popularity of restaurants in \textit{LV} and \textit{Toronto} is collected via the Google Application Programming Interfaces (API)\footnote{Due to the Google API limits and financial considerations, we could not extract popular times for all restaurants in Yelp reviews.}. For each restaurant, popular times consist of $24\times7$ entries with each entry denoting consumer flow of this restaurant during a specific hour on a specific day. The value of each entry is on a scale of 0-100 with 1 being least busy, 100 being the busiest and 0 indicating a restaurant is closed. The average daily and hourly restaurant popularity for both datasets are presented in Figure \ref{daily}-\ref{hourly}. To understand the variation of popularity for each hour across all restaurants and across the period covered by the data, we also show the standard deviation in Figure \ref{hourly}. We observe that popularity of restaurants in both cities present similar trends: on average, restaurants are most popular during lunch (i.e. 01:00 PM - 02:00 PM) and dinner (i.e. 07:00 PM - 08:00 PM)\footnote{Both shown in local time.} on Fridays and weekends. We augment the \textit{LV} and \textit{Toronto} datasets with the popularity dataset by matching restaurants' names and locations (a tuple of longitude and altitude). When this method fails or generates duplicate merges, we manually check for the correct merge. This results in two complete datasets \textit{LV} and \textit{Toronto} that include both online reviews and restaurant popularity\footnote{The data can be downloaded at \url{https://github.com/GitHubLuCheng/Effects-of-Multi-Aspect-Online-Reviews-with-Unobserved-Confounders}}. Basic statistics of both datasets are described in Table \ref{review}.

\noindent\textbf{Ethics Statement.} The Yelp data is publicly available and the Popular Times are scraped via Google API following Google's Terms of Service.
\section{Method}
We begin by illustrating our study design and rationale, and then detail the proposed framework for estimating the causal effects of multi-aspect online reviews in the presence of hidden confounders. Particularly, it consists of three stages: MAS extraction, surrogate confounder inference, and causal effect estimation.
\subsection{Study Design and Rationale}
Our research objective is to estimate the total effects and direct effects of multi-aspect online reviews on business revenue. The anchors of knowledge that we need are essentially causal. Through the causal lens, the key is to alleviate confounding biases associated with the observed effects of crowd-sourced reviews. A gold standard for unbiased estimation of causal effect is Randomized Controlled Trials (RCTs) \cite{rubin1980randomization}. However, RCTs are limited to practical use due to ethical and financial considerations. For example, it might be unethical to randomly assign consumers to write reviews for restaurants due to religious reasons (e.g., vegetarians may be assigned to barbeque restaurants). RCTs are also ungeneralizable to observational studies \cite{lopez2017estimation}. This work thereby focuses on an observational study design. Specifically, we employ a ``Consumer-Centered Model'' that uses the naturalistic self-reports of individuals regarding their dining experiences in different restaurants. As noted in Related Work, literature in various research fields provides support for using observational studies to estimate causal effects of online review systems. We acknowledge the weakness of observational studies compared to RCTs in making conclusive causal claims, however, they provide complementary advantages over RCTs in many aspects \cite{hannan2008randomized}. 

% %
\begin{figure}[ht!]
\centering
\begin{subfigure}{0.8\columnwidth}
\centering
  \includegraphics[width=\linewidth]{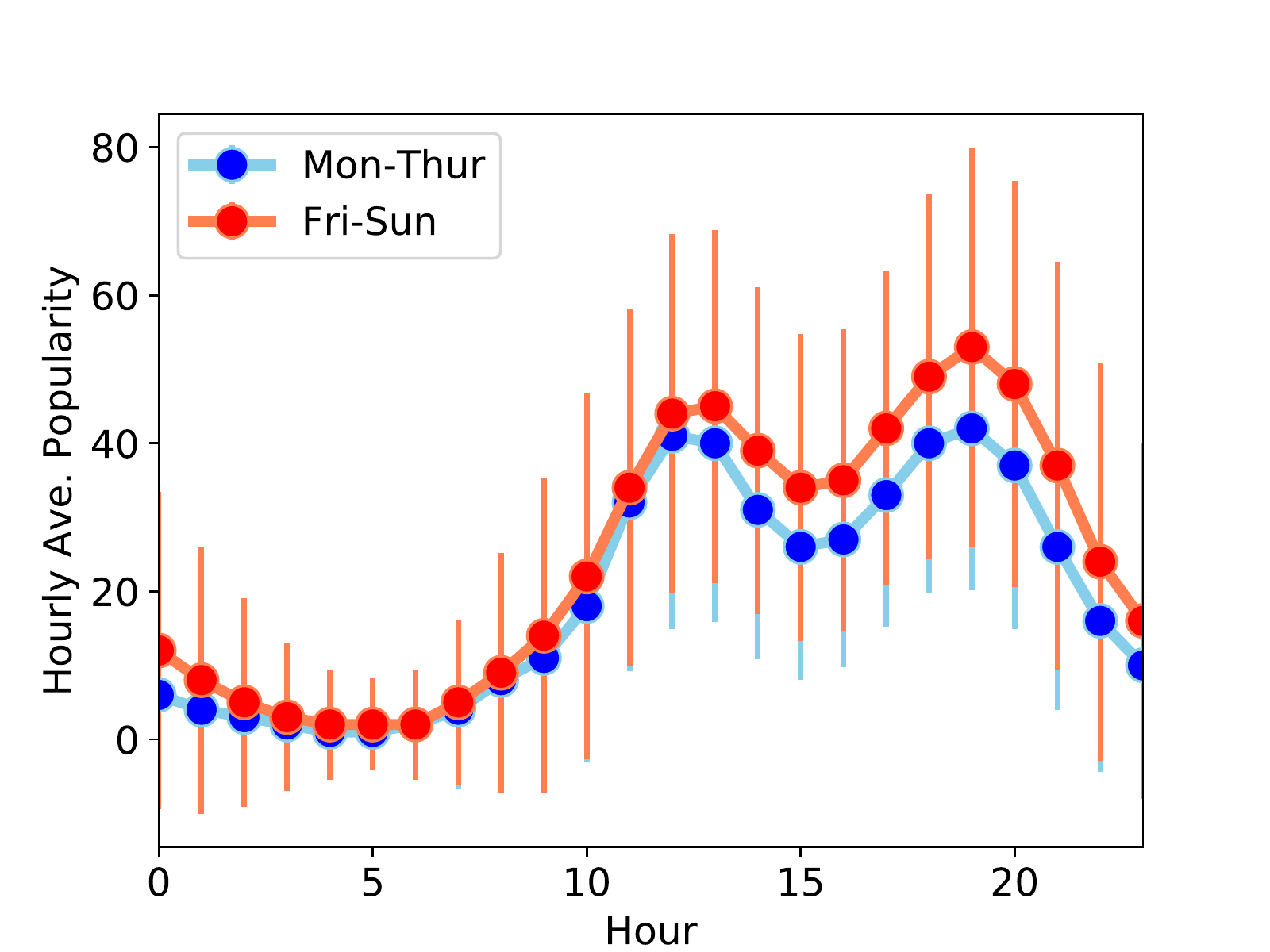}
  \caption{\textit{LV} data.}
\end{subfigure}
\begin{subfigure}{0.8\columnwidth}
\centering
  \includegraphics[width=\linewidth]{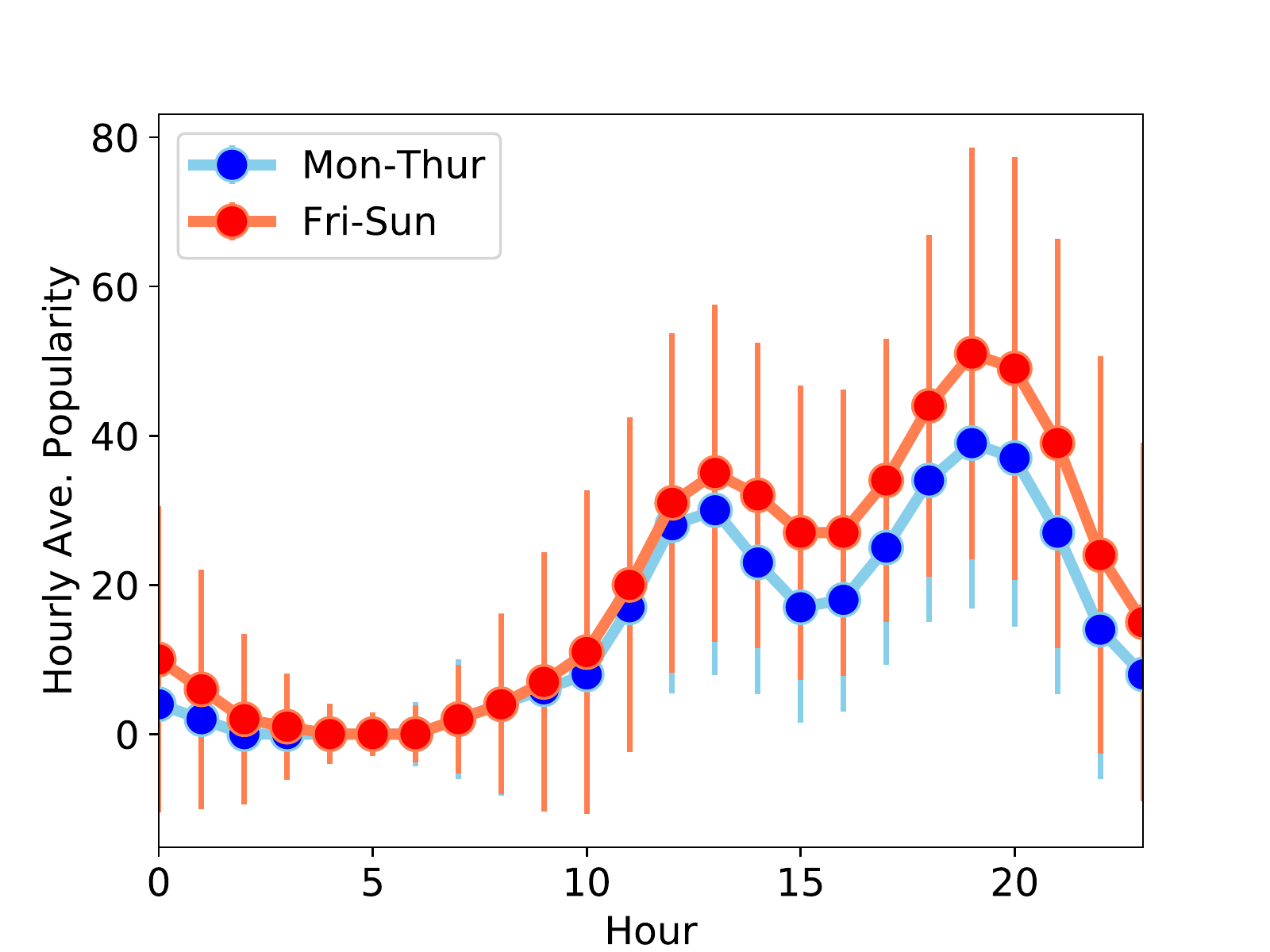}
  \caption{\textit{Toronto} data.}
\end{subfigure}
\caption{Hourly average popularity of restaurants of the day.}
 \label{hourly}
\end{figure}
This work is built under the widely recognized Potential Outcome framework \cite{rubin1980randomization} where each sentiment aspect is considered as a potential cause, ratings as the mediator (\textbf{RQ. 4}) or outcome (\textbf{RQ. 2}), and popularity as the outcome (\textbf{RQ. 1} and \textbf{RQ. 3-4}). Standard causal models (e.g., \cite{mccaffrey2013tutorial}) are inapplicable to our problem setting due to the presence of multiple continuous causes, MAS, and hidden confounders. Informed by recent advances in the intersection of machine learning and causal inference, we propose a principled framework tailored to estimating the effects of multi-aspect online reviews in the presence of hidden confounders. In \textbf{RQ. 1}, we empirically examine the validity of our framework in terms of the predictive accuracy and robust estimations of causal effects. We further answer \textbf{RQ. 2-4} by discovering the dependencies among the MAS to infer the surrogate confounders \cite{wang2019blessings}, which will be used to augment the original data. To break down the total effects of MAS into the direct and indirect effects in \textbf{RQ. 4}, we conduct a novel mediation analysis (with ratings being the mediator) by controlling for the confounding bias via the surrogate confounder. We examine whether the effects of MAS on restaurant popularity will persist after integrating numerical ratings as a mediator. We conclude with some  key theoretical implications for researchers and practical implications for businesses.
\subsection{Multi-Aspect Sentiment Extraction}
A primary challenge is to identify causes that represent typical aspects of businesses from a large corpus of textual data. A straightforward method deems each word in the Bag of Words as a cause \cite{paul2017feature}. Notwithstanding its simplicity, this method suffers from at least two limitations. Firstly, the semantic meaning of a word is highly dependent on the context and human language behavior. The estimated effect of the same word can, therefore, be inconsistent or even conflicting with each other in different reviews; secondly, words in online reviews are typically sparse and high-dimensional, which demands large computational cost and memory storage. To discover multi-dimensional causal signals from online reviews, in this work, we adopt multi-aspect sentiment analysis and focus on five widely-used aspects of restaurant reviews -- Food, Service, Price, Ambience, and Anecdotal/Miscellaneous (Misc) \cite{lu2011multi}. Our method can be extended to other aspects depending on the annotations of the training data. We detect these five aspects in each review and compute both positive and negative sentiment scores. Previous findings showed that positive and negative online reviews exert different influences \cite{tsao2019asymmetric}. 

Details of each step are described as follows: (1) \textit{Text preprocessing}. We remove the stop words, lowercase and stem the remaining words, and extract the TF-IDF representation for each review. We also employ a pre-trained neural coreference model \cite{lee2017end} to replace the pronouns in the reviews. (2) \textit{Aspect classification}. In this step, each sentence is classified into one of the five aspects.
Specifically, we segment each review into sentences and classify each sentence to an aspect using a pre-trained multi-label Na\"ive Bayes model (more details in the experimental setup).
(3) \textit{MAS computation}. We extract aspect terms and identify corresponding opinion words by cross referencing the opinion lexicon for negative and positive words\footnote{https://www.cs.uic.edu/~liub/FBS/sentiment-analysis.html}. We then assign the aspect terms to aspect categories based on the cosine similarities of word2vec using a word embedding model\footnote{https://code.google.com/archive/p/word2vec/} pretrained on the Google's News dataset\footnote{https://ai.google/tools/datasets/}. 

\begin{figure}[ht!]
\centering
\begin{subfigure}{0.8\columnwidth}
\centering
  \includegraphics[width=\linewidth]{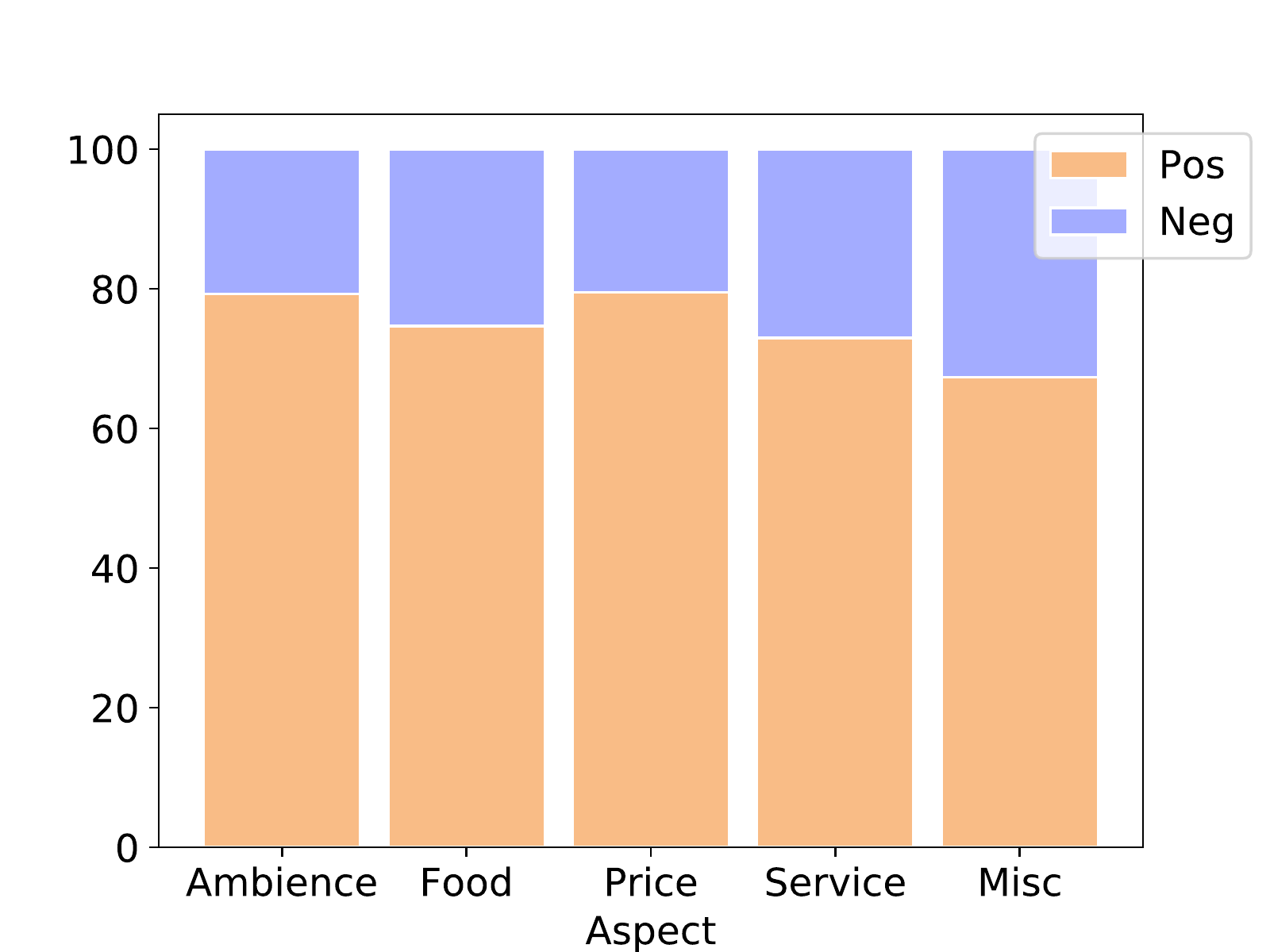}
  \caption{\textit{LV} data.}
\end{subfigure}
\begin{subfigure}{0.8\columnwidth}
\centering
  \includegraphics[width=\linewidth]{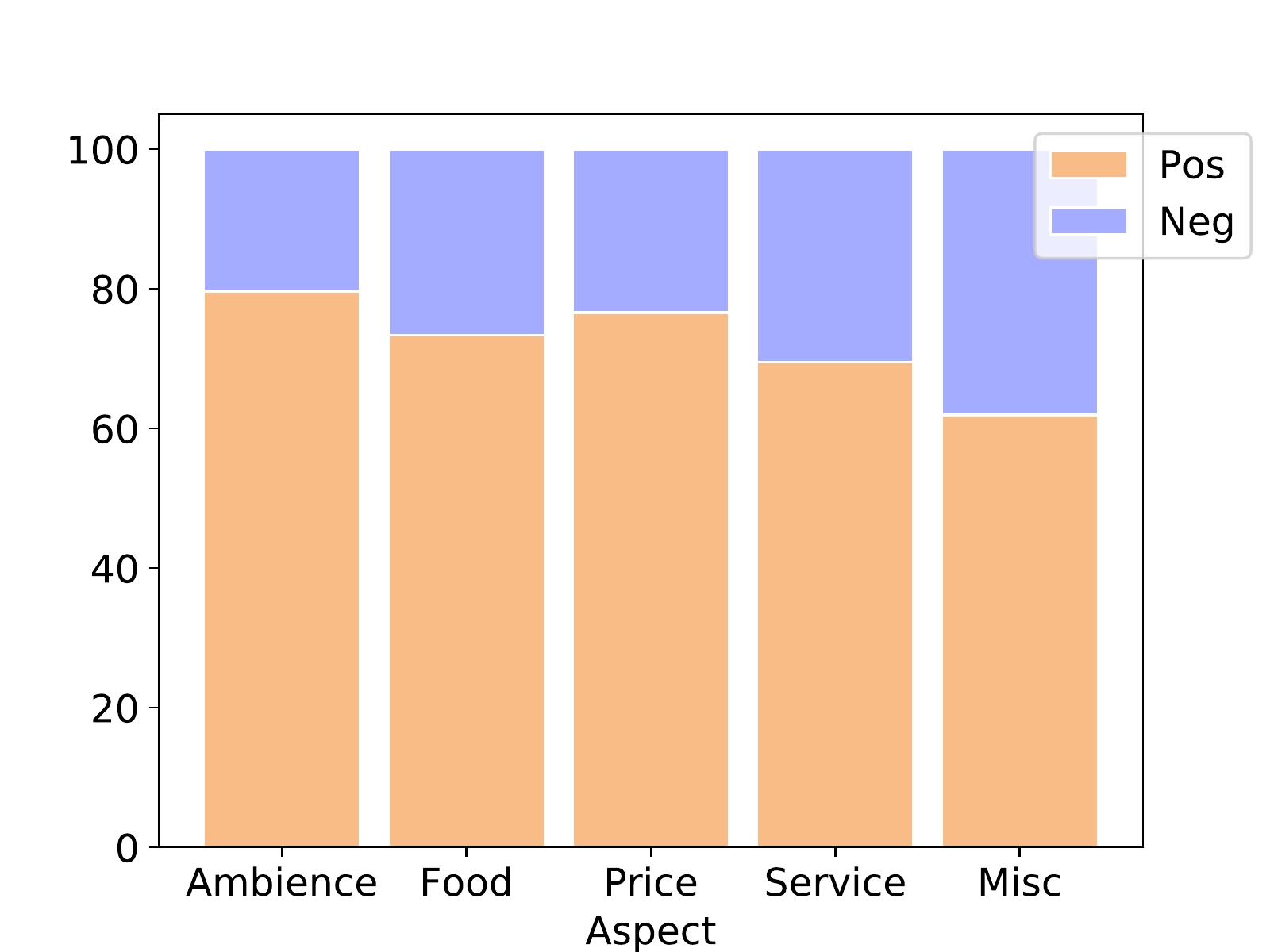}
  \caption{\textit{Toronto} data.}
\end{subfigure}
\caption{Percentages of positive and negative sentiments w.r.t. each aspect for both datasets.}
 \label{aspect_sentiment}
\end{figure}

Reviews that do not include certain aspects are treated as a Missing At Random problem \cite{little2019statistical}. That is, the missingness of aspects are not random, but might be attributed to the observed sentiment aspects, covariates of consumers and restaurants, as well as other unknown reasons \cite{rubin1976inference}. We then leverage data imputation algorithm Multivariate Imputation by Chained Equations \cite{buuren2010mice} implemented in python package ``impyute"\footnote{https://pypi.org/project/impyute/} to infer the missing values based on the existing part of the data. The final output of each review is a 10-dimensional vector with each entry being the positive and negative sentiment scores regarding each aspect. We plot the percentage of positive versus negative sentiment w.r.t. each aspect for \textit{LV} and \textit{Toronto} datasets in Figure \ref{aspect_sentiment}. As observed, results for these two datasets are similar and there are more positive reviews regarding each aspect than negative reviews.
\subsection{Surrogate Confounder Inference}
Knowing the exact nature of hidden confounders is often impossible. Therefore, we here alternatively infer a surrogate confounder from MAS (i.e., multiple causes) to mimic the properties of hidden confounders. At its core, surrogate confounder inference is a process that identifies the dependencies among MAS using unsupervised learning. This section details the process of surrogate confounder inference in multiple causal inference. 
\subsubsection{Multiple Causal Inference.} Given a corpus of $N$ reviews, each review is associated with a vector $\bm{a}$ of $m=5$ possible aspects with both positive and negative sentiments, i.e.,  $\bm{a}=(a_{1+},a_{1-},...,a_{m+},a_{m-})$, where $a_{j+}$ and $a_{j-}$ denote the positive and negative sentiment scores of the aspect $a_j$. A potential outcome function $y_i(\bm{a}): \mathbb{R}^{2m} \rightarrow \mathbb{R}$ maps configurations of these sentiment aspects to the outcome (popularity/ratings) for each restaurant $i$. Multiple causal inference seeks to characterize the sampling distribution of the potential outcomes $Y_i(\bm{a})$ for each configuration of $\bm{a}$. This distribution is essential to obtain the expected outcome for a particular array of causes $\mu(\bm{a})=\mathbb{E}[{Y_i(\bm{a)}}]$ or the average effect of an individual sentiment aspect, e.g., how much effect of textual reviews on the popularity can be attributed to the negative sentiment w.r.t. Ambience? 

Given the observational data $\mathcal{D}=\{\bm{a}_i,y_i(\bm{a}_i)\}, i\in\{1,2,...,N\}$, the fundamental problem of causal inference \cite{holland1986statistics} is we can only observe the outcome of assigned causes $y_i(\bm{a}_i)$. Without accessing to the full distribution of $Y_i(\bm{a})$ for any $\bm{a}$, a straightforward approach is to estimate conditional distribution of $\mathbb{E}[Y_i(\bm{a)}|\mat{A}_i=\bm{a}]$, where $\mat{A}_i$ is a random variable of assigned causes. Suppose we measure covariates $\mat{X}_i=\mat{x}_i$ for each restaurant (such as locations, the availability of delivery), then we have a new set of data $\mathcal{D}'=\{\bm{a}_i,\mat{x}_i, y_i(\bm{a}_i)\}, i=\{1,2,...,N\}$. Under the assumption of unconfoundedness \cite{rubin1990comment} -- covariate matrix $\mat{X}$ can capture all the confounders, we recover the full distribution of the potential outcome:
\begin{equation}
    \mathbb{E}[Y_i(\bm{a})]=\mathbb{E}[\mathbb{E}[Y_i(\bm{a})|\mat{X}_i,\mat{A}_i=\bm{a}]].
\end{equation}
\subsubsection{Surrogate Confounders.}
Classical methods for multiple causal inference assume that covariates $\mat{X}$ can fully capture the causal links between the multiple causes and the outcome, which is unverifiable in our task. To account for the hidden confounders, here, we leverage the deconfounder algorithm \cite{wang2019blessings} to infer the surrogate confounders. First, we introduce the following assumptions:
\begin{assumption} \ 
\begin{itemize}
    \item \textit{Stable Unit Treatment Value Assumption (SUTVA)} \cite{rubin1980randomization,rubin1990comment}. The SUTVA assumes that the potential outcome of one individual is independent of the assigned causes of another individual.
    \item\textit{Overlap}. The surrogate confounder $\mat{Z}_i$ satisfies:
    \begin{equation}
        p(A_{ij}\in \mathcal{A}|\mat{Z}_i)>0, \quad p(\mathcal{A})>0,
    \end{equation}
    where $A_{ij}, i=1,2...,N, j=1,2,...,2m$ is the $j$-th element of $A_i$ and $\mathcal{A}$ is the set of $A_{ij}$.
     \item \textit{No unobserved single-cause confounders}. This so-called ``single ignorability'' assumes that
    \begin{equation}
        A_{ij}\independent Y_i(\bm{a})|\mat{X}_i, \quad j=1,...,2m.
    \end{equation}
\end{itemize}
\end{assumption}
\noindent The first assumption entails that no interference exists among restaurants and there is only a single version of each sentiment aspect for every restaurant. The second assumption indicates that given the surrogate confounders, the sentiment score of at least one aspect among the five in each review is positive. The last assumption is non-standard in causal inference: there are no such hidden confounders that exclusively influence a single sentiment aspect. For example, a consumer's preferences may influence her sentiment toward both Food and Ambience. We recognize the possibility of unobserved single cause confounders, nevertheless, this requires developing more advanced method which is beyond the scope of this work. 

Next, we define and fit a latent-variable model of the assignment mechanism $p(\mat{z},a_{1+},a_{1-},...,a_{m+},a_{m-})$, where $\mat{z}\in \mat{Z}$. Specifically, the model is characterized as 
\begin{equation}
\begin{split}
     \mat{Z}_i\sim p(\cdot|\alpha)\quad i=1,...,N, \\
     A_{ij}|\mat{Z}_i\sim p(\cdot|\mat{z}_i,\theta_j)\quad j=1,...,2m,
\end{split}
\end{equation}
where $\alpha$ and $\theta_j$ are the parameters of the distribution of surrogate confounder $\mat{Z}_i$ and the per-cause distribution of $A_{ij}$, respectively. In this work, we use the latent-variable model Probabilistic PCA (PPCA) \cite{tipping1999probabilistic} following \cite{wang2019blessings}. To check if PPCA captures the population distribution of the assigned causes, we randomly hold out a subset of assigned aspects for each restaurant $i$, denoted as $\bm{a}_{i,held}$ and the rest are denoted as $\bm{a}_{i,obs}$. We then fit PPCA with $\{\bm{a}_{i,obs}\}_{i=1}^N$ and perform predictive check on the held-out dataset. A predictive check compares the observed MAS with MAS drawn from the model's predictive distribution. The predictive check score is defined as:
\begin{gather}
    p_{c}=p\big(t(\bm{a}^{rep}_{i,held})<t(\bm{a}_{i,held})\big),\\
    t(\bm{a}_{i,held})=\mathbb{E}_\mat{Z}\big[\log p(\bm{a}_{i,held}|\mat{Z})|\bm{a}_{i,obs}\big].
\end{gather}
$\bm{a}^{rep}_{i,held}$ comes from the predictive distribution:
\begin{equation}
    p(\bm{a}^{rep}_{i,held}|\bm{a}_{i,held})=\int p(\bm{a}_{i,held}|\mat{z}_i)p(\mat{z}_i|\bm{a}_{i,obs})d\mat{z}_i.
\end{equation}
Following \cite{wang2019blessings}, if the predictive check score $p_c\in(0,1)$ is larger than 0.1, we conclude that the latent-variable model can generate values of the held-out causes that give similar log likelihoods to their real values. As the threshold of 0.1 is a subjective design choice \cite{wang2019blessings}, we suggest readers referring to the original paper for more details of the predictive check score. Note that the predictive performance is not the goal but an auxiliary way of checking the model that aims to estimate causal effects \cite{shmueli2010explain,mullainathan2017machine}. We then use the fitted model $M$ to infer surrogate confounders for each restaurant, i.e., $\hat{\mat{z}}_i=\mathbb{E}_M[\mat{Z}_i|\mat{A}_i=\bm{a}_i]$. 
\subsection{Estimating Effects of MAS}
With the new input $\{\bm{a}_i,\hat{\mat{z}}_i,y_i(\bm{a}_i)\}$, we estimate the outcome model $\mathbb{E}[\mathbb{E}[Y_i(\mat{A}_i)|\mat{Z}_i=\mat{z}_i,\mat{A}_i=\bm{a}_i]]$ via simple linear regression:
\begin{equation}
    f(\bm{a},\mat{z})=\beta^T\bm{a}+\gamma^T\mat{z},
    \label{deconfounder}
\end{equation}
where $\beta$ represents a vector of the average causal effects of individual sentiment aspect and $\gamma$ is the coefficient of surrogate confounder. We now present an unbiased estimate of the causal effects of MAS \cite{wang2019blessings}:
\begin{equation}
\small
    \begin{aligned}
 \mathbb{E}_Y[Y_i(\bm{a})]-\mathbb{E}_Y[Y_i(\bm{a}')]= \\
 \mathbb{E}_{X,Z}[\mathbb{E}_Y[Y_i|\mat{A}_i=\bm{a}_i,\mat{X}_i,\mat{Z}_i]-\mathbb{E}_{X,Z}[\mathbb{E}_Y[Y_i|\mat{A}_i=\bm{a}'_i,\mat{X}_i,\mat{Z}_i].
 \label{effect}
\end{aligned}
\end{equation}

\noindent Eq. \ref{effect} assumes that the estimated effects exclusively comes from MAS, i.e., the total effects.

However, studies have shown that effects of textual reviews can be mediated by ratings \cite{li2019effect}. To further break down the total effects and examine the direct effects of MAS on restaurant popularity, we simultaneously condition on ratings -- the mediator -- and MAS. This will result in one of the two observations: 1) the effects of MAS become zero and 2) the effects of MAS change but still persist in the results. The latter indicates MAS can provide additional information about popularity that cannot be captured by numerical ratings. Therefore, we extend conventional mediation analysis framework \cite{baron1986moderator} that assumes away the presence of hidden confounders by incorporating the learned surrogate confounders into the mediation model. Note that, in this task, the surrogate confounders can only capture the pre-treatment confounding. Therefore, in addition to Assumption 1, we further assume that there is no unobserved post-treatment confounding in order to ensure the causal identification of the Deconfounder. More advanced causal models that can account for both pre-treatment and post-treatment hidden confounding will be explored in future research. 

The proposed mediation analysis consists of the following four steps:
\begin{itemize}[leftmargin=*]
    \item Step 1. Show MAS directly affects popularity.
    \item Step 2. Show MAS directly affects ratings.
    \item Step 3. Show ratings affect popularity.
    \item Step 4. Given surrogate confounders, establish a complete mediation of rating on relationship of MAS on popularity.
\end{itemize}
We accomplish the first two steps by applying the linear regression model in Eq. \ref{deconfounder}. For Step 3, we regress the popularity on ratings and conduct two-side students' t-test to check the significance of the effects. The mediation model in Step 4 is then formulated as 
\begin{equation}
    f_m(\bm{a},\mat{z},r)=\beta_m^T\bm{a}+\gamma_m^T\mat{z}+\lambda^Tr,
    \label{mediation}
\end{equation}
where $r$ is the rating and $\lambda$ is the corresponding coefficient. Lastly, we compare $\beta$ with $\beta_m$ to show the changes of the effects of MAS on restaurant popularity. Direct effect estimation is similar to Eq. \ref{effect}.
\section{Empirical Evaluation}
We conducted empirical evaluations on the two newly curated datasets to investigate the answers to our proposed research questions \textbf{RQ. 1-4}. We first delineate the experimental setup and then detail the results for each task.  
\subsection{Experimental Setup}
The experiments were implemented\footnote{Code is adapted from \url{https://colab.research.google.com/github/blei-lab/deconfounder_tutorial/}} using Tensorflow \cite{abadi2016tensorflow} and Statsmodels \cite{seabold2010statsmodels}. The dimension of the surrogate confounder $\mat{z}_i$ is set to 10 and 5 for studying the effects of MAS on ratings and restaurant popularity, respectively. The latent-variable model PPCA is optimized by Adamax \cite{kingma2014adam} with a learning rate of 0.01. In all the experiments, restaurant popularity is specified as the popularity from 07:00 PM to 08:00 PM on Saturday as it is the most popular hour within a week, as shown in Figure \ref{hourly}. Other potential forms of outcome are left to be explored in the future. 

For the validity of aspect classification described in Section 4.2, as we do not have the ground truth for the Yelp reviews in \textit{LV} and \textit{Toronto} datasets, we train a multi-label Na\"ive Bayes model on the widely used Yelp restaurant reviews with gold-standard annotations\footnote{\url{http://alt.qcri.org/semeval2014/task4/index.php?id=data-and-tools}}, a benchmark dataset in multi-aspect classification. The sample size of this dataset is 3,041, 75\% of which is used for training and the rest for testing. The multi-label Na\"ive Bayes model achieves 86.17\% accuracy in the test data. To predict MAS for the \textit{LV} and \textit{Toronto} datasets, we re-train the Na\"ive Bayes model with the entire annotated data. While we recognize there might be some differences between the distribution of the annotated data and that of our data, we believe the aspect classification on our data (\textit{LV} and \textit{Toronto}) is valid given both the annotated data and our data are from the Yelp restaurant reviews. For surrogate confounder inference, we begin by examining the correlations of all pairs of sentiment aspects, and remove highly correlated ones to ensure that the \textit{single ignorability} assumption is better satisfied. Data analysis results reveal that for both datasets, positive Ambience (Ambience Pos) are highly correlated to most of other sentiment aspects. Hence, we exclude Ambience Pos from the ten sentiment aspects.
\subsection{Results}
In this section, we present the results corresponding to the four research questions. To recall, \textbf{RQ. 1} examines the validity of the proposed framework in making causal claims; \textbf{RQ. 2-3} estimates the effects of MAS on ratings and restaurant popularity; and \textbf{RQ. 4} investigates how effects of MAS on restaurant popularity can be mediated by ratings. Unless otherwise specified, for all the results presented below, the estimated effects ($\hat{\beta}$) are shown in the column Mean, followed by the corresponding standard deviation (STD), statistical significance test, and confidence interval. We highlight all the statistically significant results. The sign of each estimated effect denotes if the effect is positive or negative.
\begin{table}[]
\centering
       
    \begin{tabular}{c|c|c|c|c}
    \hline
         Metrics& \multicolumn{2}{c|}{MSE}&\multicolumn{2}{c}{MAE} \\\hline
         \specialcell{Models}&\specialcell{Causal\\ Model}&\specialcell{Non-causal\\ Model}&\specialcell{Causal\\ Model}&\specialcell{Non-causal\\ Model}\\\hline
         \textit{LV}& \textbf{0.53}&0.55&\textbf{0.58}&0.59\\\hline
         \textit{Toronto}&\textbf{0.39}&0.40&\textbf{0.47}&0.48\\\hline
    \end{tabular}
     \caption{Predicting ratings with causal and non-causal models.}
    \label{rating_outcome}
\end{table}
\begin{table}[]
\centering
        
    \begin{tabular}{c|c|c|c|c}
    \hline
         Metrics& \multicolumn{2}{c|}{MSE}&\multicolumn{2}{c}{MAE} \\\hline
         \specialcell{Models}&\specialcell{Causal\\ Model}&\specialcell{Non-causal\\ Model}&\specialcell{Causal\\ Model}&\specialcell{Non-causal\\ Model}\\\hline
         \textit{LV}& 1.01&\textbf{0.99}&0.84&\textbf{0.83}\\\hline
         \textit{Toronto}& \textbf{0.94}&\textbf{0.94}&\textbf{0.82}&\textbf{0.82}\\\hline
    \end{tabular}
    \caption{Predicting popularity with causal and non-causal models.}
    \label{popularity_outcome}
\end{table}
\begin{table*}
\setlength\tabcolsep{1pt}
\begin{subtable}[h]{.48\textwidth}

\begin{center}
\resizebox{\textwidth}{!}{
\hfill % This is new 
\begin{tabular}{ c|c|c|c|c|c|c|c|c|c } \hline
Sentiment Aspect & $t=1$ & $t=2$ &$t=3$ & $t=4$ & $t=5$ & $t=6$ &$t=7$ & $t=8$ & $t=9$ \\ \hline\hline
Ambience Neg&-0.17&-0.18&-0.16&-0.12&-0.15&-0.14&-0.07&-0.05&-0.05\\
\rowcolor{lightgray}Food Pos& -&0.58&0.58&0.55&0.56&0.49&0.36&0.40&0.41\\
\rowcolor{lightgray}Food Neg& -& -&-0.13&-0.12&-0.11&-0.08&-0.10&-0.05&-0.05\\
\rowcolor{lightgray}Price Pos&-&-&-&0.12& 0.10&0.12&0.07&0.06&0.06\\
\rowcolor{lightgray}Price Neg&-&-&-&-&-0.07&-0.07&-0.09&-0.06&-0.05\\
\rowcolor{lightgray}Service Pos&-&-&-&-&-&0.25&0.31&0.32&0.32\\
\rowcolor{lightgray}Service Neg&-&-&-&-&-&-&-0.38&-0.35&-0.35\\
\rowcolor{lightgray}Misc Pos&-&-&-&-&-&-&-&0.17&0.17\\
Misc Neg&-&-&-&-&-&-&-&-&0.02\\\hline
\end{tabular}}
\caption{Results for causal model.}
\end{center}
\end{subtable}
 \hfill
 \begin{subtable}[h]{.48\textwidth}
\begin{center}
\resizebox{\textwidth}{!}{
\hfill % This is new 
\begin{tabular}{ c|c|c|c|c|c|c|c|c|c } \hline
Sentiment Aspect & $t=1$ & $t=2$ &$t=3$ & $t=4$ & $t=5$ & $t=6$ &$t=7$ & $t=8$ & $t=9$ \\ \hline\hline
Ambience Neg&0.05&-0.19&-0.13&-0.13&-0.12&-0.11&-0.01&-0.01&-0.01\\
\rowcolor{lightgray}Food Pos& -&0.29&0.42&0.41&0.41&0.43&0.24&0.23&0.23\\
\rowcolor{lightgray}Food Neg& -& -&-0.20&-0.20&-0.18&-0.18&-0.09&-0.08&-0.08\\
\rowcolor{lightgray}Price Pos&-&-&-&0.01& 0.05&0.04&0.04&0.03&0.03\\
\rowcolor{lightgray}Price Neg&-&-&-&-&-0.08&-0.08&-0.02&-0.03&-0.03\\
\rowcolor{lightgray}Service Pos&-&-&-&-&-&\underline{\textbf{-0.02}}&\underline{\textbf{0.22}}&0.20&0.20\\
\rowcolor{lightgray}Service Neg&-&-&-&-&-&-&-0.32&-0.31&-0.31\\
\rowcolor{lightgray}Misc Pos&-&-&-&-&-&-&-&0.03&0.03\\
Misc Neg&-&-&-&-&-&-&-&-&0.00\\\hline
\end{tabular}}
\caption{Results for non-causal model.}
\end{center}
\end{subtable}
\caption{Coefficients of causal and non-causal models predicting \textit{ratings} with sentiment aspect added one-by-one. $t=i$ indicates $i$ sentiment aspects are added into the regression models. Results with statistical significance are highlighted.}
\label{onebyone-rating}
\end{table*}
\begin{table*}
\setlength\tabcolsep{1pt}
\begin{subtable}[h]{0.48\textwidth}
\begin{center}
\hfill % This is new 
\resizebox{\textwidth}{!}{
\begin{tabular}{ c|c|c|c|c|c|c|c|c|c } \hline
Sentiment Aspect & $t=1$ & $t=2$ &$t=3$ & $t=4$ & $t=5$ & $t=6$ &$t=7$ & $t=8$ & $t=9$ \\ \hline\hline
\rowcolor{lightgray}Ambience Neg&-0.20&-0.22&-0.22&-0.23&-0.24&-0.25&-0.26&-0.26&-0.27\\
\rowcolor{lightgray}Food Pos& -&0.46&0.48&0.50&0.50&0.41&0.42&0.42&0.41\\
Food Neg& -& -&0.19&0.18&0.17&0.20&0.18&0.18&0.17\\
Price Pos&-&-&-&-0.08&-0.08&-0.04&-0.03&-0.03&-0.03\\
Price Neg&-&-&-&-&-0.04&-0.02&-0.01&-0.00&-0.01\\
Service Pos&-&-&-&-&-&0.18&0.16&0.16&0.15\\
Service Neg&-&-&-&-&-&-&0.06&0.05&0.05\\
Misc Pos&-&-&-&-&-&-&-&-0.02&-0.02\\
Misc Neg&-&-&-&-&-&-&-&-&-0.03\\\hline
\end{tabular}}
\caption{Results for causal model.}
\end{center}
\end{subtable}
 \hfill
 \begin{subtable}[h]{0.48\textwidth}
\begin{center}
\resizebox{\textwidth}{!}{
\hfill % This is new 
\begin{tabular}{ c|c|c|c|c|c|c|c|c|c } \hline
Sentiment Aspect & $t=1$ & $t=2$ &$t=3$ & $t=4$ & $t=5$ & $t=6$ &$t=7$ & $t=8$ & $t=9$ \\ \hline\hline
\rowcolor{lightgray}Ambience Neg&\underline{\textbf{0.10}}&\underline{\textbf{-0.19}}&-0.20&-0.20&-0.20&-0.21&-0.23&-0.23&-0.22\\
\rowcolor{lightgray}Food Pos& -&0.35&0.32&0.36&0.36&0.30&0.34&0.34&0.34\\
Food Neg& -& -&0.04&0.04&0.05&0.05&0.03&0.02&0.05\\
Price Pos&-&-&-&-0.04& -0.04&-0.04&-0.03&-0.02&-0.03\\
Price Neg&-&-&-&-&-0.01&-0.01&-0.03&-0.02&-0.00\\
Service Pos&-&-&-&-&-&0.07&0.02&0.04&0.03\\
Service Neg&-&-&-&-&-&-&0.06&0.06&0.09\\
Misc Pos&-&-&-&-&-&-&-&-0.02&-0.04\\
Misc Neg&-&-&-&-&-&-&-&-&-0.08\\\hline
\end{tabular}}
\caption{Results for non-causal model.}
\end{center}
\end{subtable}
\caption{Coefficients of causal and non-causal models predicting \textit{popularity} with sentiment aspect added one-by-one. $t=i$ indicates $i$ sentiment aspects are added into the regression models. Results with statistical significance are highlighted.}
\label{onebyone-popularity}
\end{table*}
\subsubsection{RQ. 1 -- Can our approach indeed make causal conclusions in contrast to non-causal models?} This task brings up the key difference between a machine learning model and a causal learning model, or, the difference between \textit{correlation} and \textit{causation}. According to the \textit{transportability theory} \cite{pearl2011transportability}, one significant difference between causal models and non-causal models, as shown in numerous works such as \cite{peters2016causal,pearl2011transportability,arjovsky2019invariant,guo2020survey}, is that the former is robust and invariant across different environments. Informed by the experimental design in \cite{wang2019blessings}, we first compare the performance of our model with that of non-causal model (both are based on simple linear regression) regarding the predictive accuracy using original data. In particular, the non-causal model directly regresses on MAS and the causal model regresses on the MAS and surrogate confounders. We then examine the robustness of the prediction results by exposing the models to various environments. We split the data into training (80\%) and test (20\%) sets and then compare the mean absolute error (MAE) and mean squared error (MSE). Results of predicting ratings and restaurant popularity using original data are presented in Table \ref{rating_outcome}-\ref{popularity_outcome}. We first observe that incorporating hidden confounders does not exacerbate the predictive accuracy, but rather shows competitive performance compared to non-causal model. 

Next, we show the robustness of our model by adding the sentiment aspect into the outcome model Eq. \ref{deconfounder} \textit{one by one}, as suggested by \cite{wang2019blessings}. We then examine whether the signs of the coefficients flip or not while predicting the ratings and popularity. A causal model is expected to output coefficients with consistent signs when more sentiment aspects are included into the system whereas a non-causal model may output coefficients with inconsistent signs \cite{wang2019blessings}. We use \textit{Toronto} dataset as an example as similar results can be found using \textit{LV} dataset. We here focus on coefficients with statistical significance (highlighted in grey) and report results in Table \ref{onebyone-rating}-\ref{onebyone-popularity} (coefficients with flipped signs are highlighted in bold font). We observe that coefficients of non-causal models flip the signs whereas those of causal models do not change as we include more sentiment aspects. For example, in the task of predicting ratings, the coefficient of Service Pos in the non-causal model is negative with 6 sentiment aspects included in the system but changes to positive when we add the 7-th sentiment aspect. This suggests that our approach indeed controls for the confounders and can obtain more causality-driven results compared to non-causal models. 
\subsubsection{RQ. 2 -- Effects of MAS on Ratings}
In this task, the predictive check scores (Eq. 5) for surrogate confounder inference are 0.78 and 0.85 (both are larger than 0.1) for \textit{LV} and \textit{Toronto}, respectively. The estimated effects of MAS on ratings can be seen in Table \ref{As_rating_LV}-\ref{As_rating_To}. 

For the \textit{LV} dataset, causal effects of the negative sentiment regarding Ambience, Food, Price, Service and Anecdotal (Misc), and the positive sentiment w.r.t. Service are statistically significant in terms of their influence on the ratings.
Similarly, positive Misc reviews as well as both positive and negative reviews regarding Food, Price and Service have statistically significant causal effects on Yelp ratings for the \textit{Toronto} dataset.
We also observe that sentiment w.r.t. Service and Food have the largest and the second largest effect size. Of particular interest is that, for the \textit{LV} dataset, negative sentiments w.r.t. various aspects tend to have stronger influence on the ratings than positive sentiments. For example, effect size of negative Service review (0.60) is 131\% larger than that of positive Service review (0.26); In contrast, for the \textit{Toronto} dataset, positive sentiment tends to have larger influence on the ratings.
Results for both datasets show larger influence of negative Service reviews than that of positive Service reviews (68\% larger for \textit{Toronto} dataset). Another observation is that sentiment aspects that have significant effects are mostly negative for \textit{LV} whereas for \textit{Toronto} dataset, both positive and negative MAS significantly influence the Yelp ratings.
\begin{table}
\setlength\tabcolsep{3pt}
\begin{center}
\hfil % This is new 
\begin{tabular}{ c|c|c|c|c|c } \hline
Sentiment Aspect & Mean & STD &$p$-value & [0.025 & 0.975] \\ \hline\hline
Intercept&3.47 &0.02 & $0.00^*$ & 3.44&3.50 \\
\rowcolor{lightgray} Ambience Neg&-0.12&0.05&$0.01^*$&-0.22&-0.02\\
Food Pos& 0.02&0.07&0.78&-0.12&0.17\\
\rowcolor{lightgray}Food Neg& -0.23& 0.06&$0.00^*$&-0.34&-0.11\\
Price Pos&0.05&0.05&0.32&-0.05&0.14\\
\rowcolor{lightgray}Price Neg&-0.12&0.06&$0.05^*$&-0.25&-0.00\\
\rowcolor{lightgray}Service Pos&0.26&0.07&$0.00^*$&0.13&0.40\\
\rowcolor{lightgray}Service Neg&-0.60&0.05&$0.00^*$&-0.71&-0.50\\
Misc Pos&-0.06&0.07&0.38&-0.19&0.07\\
\rowcolor{lightgray}Misc Neg&-0.13&0.04&$0.00^*$&-0.21&-0.05\\\hline
\end{tabular}
\caption{Effects of MAS on ratings for \textit{LV} dataset\textsuperscript{1}.}
{\small\textsuperscript{1} $^*$ denotes 5\% significance and $^{**}$ 10\% significance.}
\label{As_rating_LV}
\end{center}
\end{table}
\begin{table}
\setlength\tabcolsep{3pt}
\begin{center}
\hfil % This is new 
\begin{tabular}{ c|c|c|c|c|c } \hline
Aspect Sentiments & Mean & STD &$p$-value & [0.025 & 0.975] \\ \hline\hline
Intercept&3.49 &0.01 & $0.00^*$ & 3.47&3.51 \\
Ambience Neg&-0.02&0.03&0.41&-0.08&0.03\\
\rowcolor{lightgray}Food Pos& 0.25&0.05&$0.00^*$&0.15&0.34\\
\rowcolor{lightgray}Food Neg& -0.06& 0.04&$0.10^{**}$&-0.14&0.01\\
\rowcolor{lightgray}Price Pos&0.06&0.03&$0.07^{**}$&-0.01&0.12\\
\rowcolor{lightgray}Price Neg&-0.05&0.03&$0.08^{**}$&-0.11&0.01\\
\rowcolor{lightgray}Service Pos&0.22&0.05&$0.00^*$&0.13&0.31\\
\rowcolor{lightgray}Service Neg&-0.37&0.03&$0.00^*$&-0.44&-0.30\\
\rowcolor{lightgray}Misc Pos&0.05&0.03&$0.09^{**}$&-0.01&0.12\\
Misc Neg&-0.03&0.03&0.28&-0.09&0.03\\\hline
\end{tabular}
\caption{Effects of MAS on ratings for \textit{Toronto} dataset.}
\label{As_rating_To}
\end{center}
\end{table}
\subsubsection{RQ. 3 -- Effects of MAS on Restaurant Popularity}
The predictive check scores of the surrogate inference model in the second task are 0.78 and 0.87 for \textit{Toronto} and \textit{LV}, respectively. We present the results in Table \ref{As_popularity_LV}-\ref{As_popularity_To}. 

Compared to effects on Yelp ratings, fewer sentiment aspects have statistically significant effects on restaurant popularity. In particular, negative reviews regarding Food and positive reviews regarding Service are found causally related to popularity for \textit{LV} dataset. For \textit{Toronto}, the identified causes are negative sentiment regarding Ambience and positive sentiment regarding Food. We also observe that the effect size of Service Pos (0.29) and Food Neg (0.26) are similar for \textit{LV} whereas the effect size of Food Pos (0.39) is relatively larger than that of Ambience Neg (0.24) for \textit{Toronto}.
%%%%%%%%%%%%%
\subsubsection{RQ. 4 -- Direct Effect of MAS on Restaurant Popularity}
The predictive check scores of the mediation model are 0.87 and 0.77 for \textit{LV} and \textit{Toronto} datasets, respectively. The effect of Yelp ratings on popularity at Step 3 in the mediation model is also found statistically significant. In this task, we compare the total effects of MAS with its direct effect on popularity. Only the results that are statistically significant are presented in Table \ref{media_res}. The third ($\hat{\beta}$) and fourth ($\hat{\beta}_m$) rows denote the estimated effects of MAS on popularity before and after integrating the mediator ratings. We begin by noticing that effects from textual reviews that carry negative and positive aspects of the restaurants persist in the mediation model. As expected, ratings slightly reduce the effect size of MAS regarding both positive and negative sentiment aspects. In particular, ratings cancel out the causal effects of MAS on popularity such that the effect size of both negative and positive sentiment aspects are driven towards zero. The conclusions apply to both datasets. 

\begin{table}
\setlength\tabcolsep{3pt}
\begin{center}
\hfil % This is new 
\begin{tabular}{ c|c|c|c|c|c } \hline
Sentiment Aspect & Mean & STD &$p$-value & [0.025 & 0.975] \\ \hline\hline
Intercept&0.03 &0.03 & 0.34 & -0.02&0.07 \\
Ambience Neg&-0.10&0.08&0.20&-0.22&0.11\\
Food Pos& 0.11&0.11&$0.32$&-0.04&0.35\\
\rowcolor{lightgray}Food Neg& -0.26& 0.10&$0.01^*$&-0.32&0.00\\
Price Pos&-0.10&0.08&0.22&-0.18&0.09\\
Price Neg&-0.03&0.08&0.76&-0.13&0.16\\
\rowcolor{lightgray}Service Pos&0.29&0.10&$0.01^*$&0.20&0.61\\
Service Neg&-0.06&0.08&0.46&-0.06&0.18\\
Misc Pos&-0.14&0.09&0.13&-0.17&0.14\\
Misc Neg&-0.03&0.05&0.63&-0.01&0.21\\\hline
\end{tabular}
\caption{Effects of MAS on Popularity for \textit{LV} dataset.}
\label{As_popularity_LV}
\end{center}
\end{table}
\begin{table}
\setlength\tabcolsep{3pt}
\begin{center}
\hfil % This is new 
\begin{tabular}{ c|c|c|c|c|c } \hline
Sentiment Aspect & Mean & STD &$p$-value & [0.025 & 0.975] \\ \hline\hline
Intercept&-0.03 &0.03 & 0.37 & -0.05&0.03 \\
\rowcolor{lightgray}Ambience Neg&-0.24&0.06&$0.00^*$&-0.35&-0.12\\
\rowcolor{lightgray}Food Pos& 0.39&0.08&$0.00^*$&0.15&0.49\\
Food Neg& 0.09& 0.08&0.29&-0.09&0.14\\
Price Pos&-0.04&0.06&0.48&-0.17&0.04\\
Price Neg&0.03&0.05&0.60&-0.12&0.08\\
Service Pos&0.10&0.08&0.22&-0.15&0.16\\
Service Neg&0.08&0.06&0.16&-0.04&0.19\\
Misc Pos&0.03&0.06&0.62&-0.16&0.09\\
Misc Neg&-0.03&0.06&0.64&-0.21&0.02\\\hline
\end{tabular}
\caption{Effects of MAS on Popularity for \textit{Toronto} dataset.}
\label{As_popularity_To}
\end{center}
\end{table}
\textbf{In summary}, our answers to \textbf{RQ. 1-4} show that (1) our framework can control for hidden confounders and identify causality-driven effects of multi-aspect online reviews; (2) most of sentiment aspects are found causally related to ratings and the effects are different; (3) only a few sentiment aspects are found causally related to popularity and those effects are different; and (4) the mediator, numerical ratings, can cancel out the effects of MAS on popularity.
\section{Implication}
This study examines the causal effects of multi-aspect textual reviews on ratings and business revenue (reflected by popularity) using observational data. Our work presents several compelling contributions: (i) In contrast to single-cause-based causal effect estimation, we propose to differentiate the effects of textual reviews from multiple dimensions; (ii) in addition to the total effect, we also investigate the direct effects of textual reviews and show that they can indeed provide additional information besides numerical ratings; and (iii) due to the common presence of hidden confounders in observational studies, we employ the advanced causal learning models to control for the hidden confounding biases. Findings drawn from empirical evaluations on two newly curated datasets show that our approach can help restaurateurs strategize business operations by focusing on those aspects that are more relevant to business revenue. We illustrate the implications of our contributions in the remainder of this section. \textit{The discussions below are not intended to bring up strategical plans that can solve problems for the entire restaurant businesses, but rather showcase the specific solutions to restaurants in well-directed scenarios. Our proposed approach can be easily adapted to new scenarios. }
\begin{table}[]
\centering
    \begin{tabular}{c|c|c|c|c}
    \hline
         Datasets& \multicolumn{2}{c|}{\textit{LV}}&\multicolumn{2}{c}{\textit{Toronto}} \\\hline
         \specialcell{Aspects \\Sentiment}&\specialcell{Food\\ Neg}&\specialcell{Service\\ Pos}&\specialcell{ Ambience \\ Neg}&\specialcell{Food \\Pos}\\\hline
         $\hat{\beta}$& $-0.26^*$&$0.29^*$&$-0.24^*$&$0.39^*$\\\hline
         $\hat{\beta}_m$&$-0.25^*$&$0.23^*$&$-0.22^*$&$0.35^*$\\\hline
    \end{tabular}
        \caption{Results of mediation analysis for both datasets.}
    \label{media_res}
\end{table}
\subsection{Insights about Improving Ratings}
Empirical results from Table \ref{As_rating_LV}-\ref{As_rating_To} suggest that for the positive and negative reviews w.r.t. each aspect of restaurant, the restaurateurs should use different operation strategies to improve these aspects. Effects of positive and negative aspects on restaurant ratings can be significantly different. This agrees on previous studies on asymmetric effects of positive and negative sentiments \cite{tsao2019asymmetric}. Moreover, these findings vary across cities. Our research indicates that consumers in Las Vegas have stronger tendency to write negative reviews w.r.t. different aspects than consumers in Toronto. We conjecture that 1) as Las Vegas is an internationally renowned major resort city and is known primarily for fine dining and entertainment, it has much larger floating population such as tourists and leisure travelers. One primary difference between tourists and local residents is they may have higher expectations to food and service of the restaurants; and 2) consumers in Las Vegas may have more serious considerations for reviews and recommendations because they are more likely to use online review systems to read historical reviews and write new reviews for future consumers. Consequently, for restaurants in Las Vegas, our study suggests restaurateurs \textit{largely improving consumer service} and \textit{avoiding negative reviews w.r.t. other restaurant aspects}. For restaurants in Toronto, our empirical results imply that restaurateurs might first focus on \textit{improving both consumer service and food quality.}
\subsection{Insights about Gaining Popularity}
Direct effects of different sentiment aspects persisting in the results implies that in addition to ratings, it is important for restaurateurs to understand the effects of multi-aspect textual reviews to gain popularity more effectively. Drawing on the experimental results in Table \ref{As_popularity_LV}-\ref{As_popularity_To}, we also conclude that restaurant popularity is causally affected by a few primary aspects, namely, Food, Service, and Ambience. In particular, to improve popularity, our study suggests that restaurateurs in Las Vegas \textit{reduce number of negative reviews regarding food} meanwhile \textit{largely improve consumer service}. Restaurateurs in Toronto might \textit{avoid negative reviews regarding restaurants' ambience} and \textit{largely improve food quality}. 

There are fewer sentiment aspects found statistically significant compared to the results for ratings. This is mainly because there are potentially many other factors besides reviews that can influence restaurant popularity, such as the locations and price ranges of restaurants. For example, restaurants that are closer to populous places (e.g., Time Square in New York) may have larger consumer flow than restaurants closer to residential areas. Fast food can be popular due to its convenience and low price range. Bars are often most popular at night, and restaurants for breakfast and brunch are most popular before noon. Therefore, popularity is a much more complex and ephemeral measure \cite{trattner2018predictability} and our study suggests promising research directions to explore in the future. In this work, we interpret our estimates as \textit{lower bounds} of the effects of online review systems on popularity. Other potential factors to consider include business locations, price range, categories, photos uploaded in online reviews, temporal confounders, and the review herding effects.
\section{Discussions}
We identify how multi-aspect online reviews can facilitate more nuanced understandings than single numerical causes such as ratings and provide unique perspectives to the business operation and strategies. To achieve this, we propose three novel research questions to examine the causal relations among MAS, ratings, and business popularity. To address the defining challenge in causal inference -- confounding -- we employ a multiple-causal-inference framework with hidden confounders and leverage the advanced techniques in causal learning to control for the confounding biases. Empirical results on two novel datasets corroborate the importance of multi-aspect online reviews in shaping business operation and strategies in terms of different sentiment aspects. Our work attests further research in this new space and opens up intriguing opportunities beyond existing reporting methodologies. 

This study is not without limitations. First, our results are likely to be influenced by \textit{selection bias} in consumers who choose to publicly self-report their dining experiences on Yelp. There is also inherent issues of selection bias in who is on Yelp and the differences between various online review systems. It is imperative not to take the datasets as being representative of the countries we study, or individuals included in the datasets. In order to provide more general advice for restaurateurs, experiments on various datasets w.r.t. different cities need to be conducted. Second, there might be potential biases introduced by using the off-the-shelf approaches for MAS extraction such as data bias and algorithmic bias. How to alleviate the biases in MAS extraction is critical to obtain more valid causal effects estimation of multi-aspect online reviews. This might be compensated by human-in-the-loop validation of the MAS extraction results. 

Third, while we focus on numerical ratings and textual reviews, we recognize the fact that many other factors such as photos posted by consumers, restaurants' categories, locations, price ranges and the availability of delivery can influence restaurant popularity. Further, future works can adopt methods such as location-based segmentation to better account for geo-related confounders. Considering that potential confounding factors can 
have temporal dynamics (e.g., consumers' preferences to food type) and there might be review herding effects (e.g., a consumer's review can be affected by historical reviews such that ``rich gets richer''), we need to address these more complex scenarios with advanced causal approaches in future work.

Our work can also be improved by complementary offline information (e.g., reviews from professional reviewers) and investigation of the authenticity of Yelp reviews. Previous work \cite{anderson2012learning} revealed restaurateurs' strong incentives to leave fake positive reviews to combat new incoming reviews. Consequently, future direction can also be directed toward detecting fake reviews and controlling their influence.
Although the overall research framework can be applied to other domains (e.g., book sales), caution is warranted when generalizing the specific findings to other domains. Future research may be conducted to test the hypothesis in other domains. Another future research of this work is to extend the Deconfounder or develop more advanced causal models to tackle the causal mediation analysis and selection bias problems in a multiple causal inference setting.
We also acknowledge that there have been discussions, e.g., \citep{ogburn2019comment,imai2019comment}, about the identification issues with Deconfounder, such as the sufficiency of the conditional independence assumptions to formalize ``single ignorability'' assumption \cite{imai2019comment}.
Therefore, when the required assumptions of Deconfounder are violated and its causal identification is not guaranteed, a rigorous theoretical analysis of the algorithm and sensitivity analysis of the experimental results are needed to help understand the robustness of the empirical findings.

\section*{Acknowledgements}
This material is based upon work supported by, or in part by, the U.S. Office of Naval Research (ONR) and the U.S. Army Materiel Command (AMC) under contract/grant number N00014-21-1-4002, W911NF2110030, and by ARL under grant W911NF2020124, as well as the National Science Foundation (NSF) under grant numbers 2125246, 1633381, and 1610282. We thank Dr. Kai Shu for his invaluable suggestions.
\bibliography{aaai21}
\end{document}